\documentclass[]{spie}

\usepackage{amsmath,amsfonts,amssymb}
\usepackage{hhline}
\usepackage{arydshln} 
\usepackage{graphicx}
\usepackage[colorlinks=true, allcolors=blue]{hyperref}
\usepackage{comment}
\usepackage[skip=0.5pt]{caption}

\title{A Hybrid CNN and ML Framework for Multi-modal Classification of Movement Disorders Using MRI and Brain Structural Features}

\author[a]{Mengyu Li}
\author[b]{Ingibjörg Kristjánsdóttir}
\author[c,d]{Thilo van Eimeren}
\author[c,d]{Kathrin Giehl}
\author[a]{Lotta M. Ellingsen}
\author[*]{the ASAP Neuroimaging Initiative}

\affil[a]{University of Iceland, Faculty of Electrical and Computer Engineering, Reykjavik, Iceland}
\affil[b]{University of Iceland, Faculty of Medicine, Reykjavik, Iceland}
\affil[c]{University of Cologne, Faculty of Medicine, Cologne, Germany}
\affil[d]{University Hospital Cologne, Dept. of Nuclear Medicine and Dept. of Neurology, Cologne, Germany}

\authorinfo{Further author information: Send correspondence to M. Li; E-mail: mel17@hi.is}

\begin{document} 
\maketitle
\begin{abstract}
Atypical Parkinsonian Disorders (APD), also known as Parkinson-plus syndrome, are a group of neurodegenerative diseases that include progressive supranuclear palsy (PSP) and multiple system atrophy (MSA). In the early stages, overlapping clinical features often lead to misdiagnosis as Parkinson’s disease (PD). Identifying reliable imaging biomarkers for early differential diagnosis remains a critical challenge. In this study, we propose a hybrid framework combining convolutional neural networks (CNNs) with machine learning (ML) techniques to classify APD subtypes versus PD and distinguish between the subtypes themselves: PSP vs. PD, MSA vs. PD, and PSP vs. MSA. The model leverages multi-modal input data, including T1-weighted magnetic resonance imaging (MRI), segmentation masks of 12 deep brain structures associated with APD, and their corresponding volumetric measurements. By integrating these complementary modalities, including image data, structural segmentation masks, and quantitative volume features, the hybrid approach achieved promising classification performance with area under the curve (AUC) scores of 0.95 for PSP vs. PD, 0.86 for MSA vs. PD, and 0.92 for PSP vs. MSA. These results highlight the potential of combining spatial and structural information for robust subtype differentiation. In conclusion, this study demonstrates that fusing CNN-based image features with volume-based ML inputs improves classification accuracy for APD subtypes. The proposed approach may contribute to more reliable early-stage diagnosis, facilitating timely and targeted interventions in clinical practice.
\end{abstract}


\section{Introduction}
Atypical Parkinsonian Disorders (APD), including progressive supranuclear palsy (PSP) and multiple system atrophy (MSA), are neurodegenerative disorders that share overlapping clinical features with Parkinson’s disease (PD), particularly in the early stages. This often results in delayed or incorrect diagnosis, underscoring the need for accurate and robust biomarkers for early differentiation. PSP and MSA are prototypical examples of diseases that often present with very specific structural characteristics in the brain. These include atrophy in deep-brain structures, such as the brainstem in PSP patients and the striatum and the brainstem in MSA patients. Magnetic resonance imaging (MRI) plays an important role in distinguishing APD from PD and from each other.

Prior studies have focused primarily on morphological features derived from MRI, such as brainstem atrophy. For example, Silsby et al. and Constantinides et al. manually measured midbrain and pons areas from mid-sagittal slices to compute the midbrain to pons (M/P) area ratio and the Magnetic Resonance Parkinsonism Index (MRPI), which has been shown to have a discriminative power between PSP and controls.~\cite{Silsby2017, Constantinides2018} However, these approaches rely on 2D measurements and manual delineations, limiting scalability and comprehensiveness. Sjöström et al. used FreeSurfer~\cite{iglesias2015} to automatically segment the substructures of the brainstem for volumetric analysis in the differential diagnosis of PSP from PD and MSA.~\cite{SJOSTROM202018} However, this analysis was limited to brainstem substructures only.  

The advent of Deep Learning (DL) has catalyzed a paradigm shift from manual biometry to automated feature discovery. Kiryu et al. provided a foundational proof-of-concept, demonstrating that Convolutional Neural Networks (CNNs) trained on single midsagittal(2D) T1-weighted slices could achieve diagnostic accuracy exceeding 90\% for differentiating PD, PSP, and MSA.~\cite{Kiryu} Building on this, recent advancements have transitioned towards 3D volumetric analysis to capture more subtle, diffuse microstructural changes. Desai et al. illustrated the potential of this approach by employing 3D DenseNet architectures combined with evolutionary algorithms for feature selection, achieving high sensitivity in distinguishing PD from controls.~\cite{DESAI} Similarly, Volkmann et al. demonstrated the efficacy of machine learning in classifying between PD and PSP using raw imaging data, such as T1-weighted MRI and diffusion tensor imaging (DTI).~\cite{Volkmann2025} 

However, a critical limitation persists in many state-of-the-art deep learning models, which is the "black box" phenomenon. High-dimensional models, such as those employing complex 3D CNNs, often lack interpretability, making it difficult for clinicians to verify whether the classification is driven by pathological tissue changes or irrelevant background noise. Recent trends have thus pivoted towards Explainable AI (XAI) and segmentation-guided classification. For instance, Ling et al. introduced a radiomics-guided deep learning framework that constrains the neural network using interpretable biological features, significantly enhancing both performance and trust.~\cite{Ling} Furthermore, the prospective Automated Imaging Differentiation for Parkinsonism (AIDP) study (Vaillancourt et al.) has underscored the clinical necessity of validating these automated markers in multi-center settings.~\cite{Vaillancourt} Yang et al.~\cite{yang2018} adapted Gradient-weighted Class Activation Mapping (Grad-CAM)~\cite{gradcam} to 3D neuroimaging, demonstrating its utility in localizing hippocampal atrophy in Alzheimer’s disease.

Addressing the dual challenge of diagnostic accuracy and interpretability, this study proposes a hybrid framework that combines convolutional neural network (CNN) and traditional machine learning (ML) models to classify APD subtypes based on T1-weighted MRI scans and volumetric data with structural segmentation guidance. Unlike approaches that rely solely on global volumetric features or raw image intensities, we integrate multi-modal inputs comprising raw MRI data, explicit segmentation masks of 12 deep brain structures, and their corresponding volumetric measurements. Using a large clinical cohort of T1-weighted MRI scans, including PD, PSP, and MSA subjects, our experiments on PSP vs. PD, MSA vs. PD, and PSP vs. MSA classification tasks demonstrate that incorporating segmentation masks alongside MRI and volume data improves classification performance over using volume or MRI alone. This segmentation-guided approach acts as a hard attention mechanism, forcing the network to prioritize anatomically relevant regions. Furthermore, we employ 3D Grad-CAM and generate resultant attention maps to visualize the model’s voxel-wise focus, providing direct evidence that the classifier attends to disease-specific pathologies, bridging the gap between advanced deep learning and clinical interpretability.

\section{METHODS}
\subsection{Data}
This study utilized a multi-site clinical MRI dataset comprising T1-weighted scans of 554 subjects diagnosed with various movement disorders. Curated data were obtained from the curated ASAP Neuroimaging Initiative (see Acknowledgements). The cohort was categorized into five groups: 285 PD, 192 PSP, 23 unclassified MSA, 20 MSA-Cerebellar type (MSA-C), and 34 MSA-Parkinsonian type (MSA-P). Each subject's image was processed using a region-based segmentation pipeline~\cite{Li2025}, which generated segmentation masks for 12 deep brain structures relevant to APD. These included substructures within the brainstem (midbrain, pons, medulla oblongata, and superior cerebellar peduncles), the ventricular system (left and right lateral ventricles, and the 3rd, and 4th ventricles), and striatum structures (left and right putamen and caudate nuclei). 

Based on the segmentation results, the three-dimensional (3D) volume of each of the 12 structures was computed and further normalized by intracranial volume (ICV) to account for inter-subject variability. The resulting dataset for the classifier comprised three data modalities per subject: T1-weighted MRI, binary segmentation masks for 12 regions, and 12-dimensional tabular vector of structural volumes. These multi-modal inputs were used for downstream classification tasks among APD subtypes (PSP vs. PD, MSA vs. PD, and PSP vs. MSA).

\subsection{Hybrid framework}
\begin{figure}[!htbp]
\centering
\includegraphics[width=0.8\textwidth]{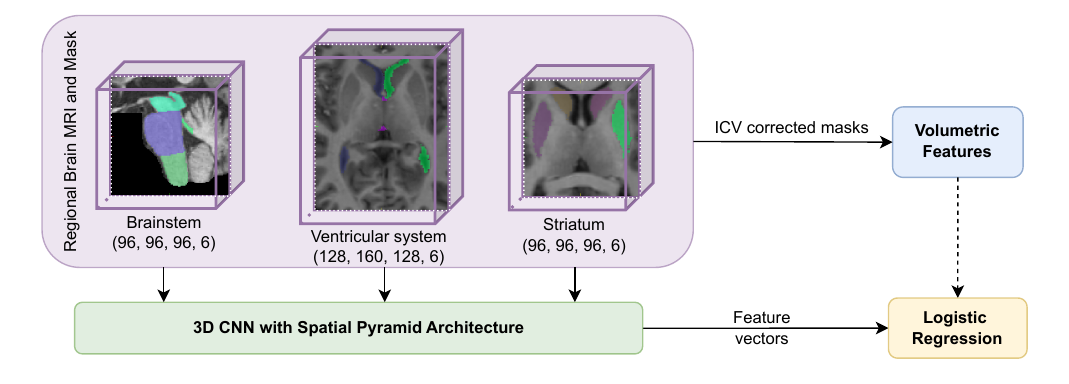}
\caption{The workflow of the proposed hybrid CNN and ML classification framework}
\label{fig:framework}
\end{figure}

We propose a hybrid classification framework that integrates deep learning-based image features with structure-specific volumetric data using a two-stage pipeline (see Figure~\ref{fig:framework}). In the first stage, we train a 3D CNN with a spatial pyramid-inspired architecture~\cite{He2014} to extract high-level features from multi-regional brain MRI and the corresponding segmentation mask inputs. The model consists of three input branches, each corresponding to a brain region of interest: the brainstem, ventricles, and striatum. Each branch is composed of three Conv3D-BatchNorm-Pooling blocks followed by a global average pooling layer. The extracted features from each branch are concatenated and passed through a dense layer with 256 units and dropout regularization before final binary classification using a sigmoid activation. In the second stage, we extract the dense-layer feature vectors from the trained CNN and concatenate them with 12 normalized volumetric measurements derived from the segmentation masks. The fused feature vectors are then input into a logistic regression classifier with L2 regularization and class-balanced weights. This architecture allows for the integration of both voxel-level texture and region-level structural information, leveraging their complementary strengths for improved disease subtype classification. 

\subsection{Training, validation, and testing}
We split the full dataset into 80\% for training and validation and 20\% for final, held-out testing. The 3D CNN with a spatial pyramid-inspired architecture was trained on the 80\% subset using five-fold stratified cross-validation to monitor training dynamics and ensure robustness. For the final model used in downstream fusion, the CNN was retrained on the entire 80\% training/validation set to avoid validation-induced selection bias. The CNN was trained with a batch size of 2, using the Adam optimizer (learning rate = $1 \times 10^{-4}$) and a weighted binary cross-entropy loss to address class imbalance. Early stopping and learning rate reduction on plateau were employed to stabilize training. After training, features from the final dense layer were extracted. These CNN-derived features were concatenated with 12 ICV-corrected volumetric features and used to train a logistic regression classifier with L2 regularization and class-balanced weighting. The performance of the final hybrid model was evaluated on the held-out test set using standard classification metrics, including accuracy, Youden’s index, F1 score, AUC-Receiver Operating Characteristic (ROC), sensitivity, and specificity. 

\subsection{Discriminative Region Visualization via 3D Grad-CAM}
To validate the interpretability of our hybrid framework, we employed 3D Gradient-weighted Class Activation Mapping (Grad-CAM)~\cite{gradcam} to visualize the specific anatomical sub-regions driving the classification decisions. Unlike traditional 2D approaches, our implementation computes class-specific gradients with respect to the volumetric feature maps of the final convolutional layer for each anatomical branch. For a given class $c$ and a feature map $A^k$ in the target layer, the neuron importance weights $\alpha_k^c$ were calculated via global average pooling of the gradients. The weighted combination of these feature maps was processed through a Rectified Linear Unit (ReLU) to isolate positive contributions to the target class:

\begin{equation}
L^{c}_{\text{Grad-CAM}} = \mathrm{ReLU}\!\left( \sum_{k} \alpha^{c}_{k} A^{k} \right)
\end{equation}

The resulting low-resolution 3D activation maps were upsampled to the original MRI resolution using cubic interpolation and overlaid onto the subject's T1-weighted scan. To capture pathological patterns that are robust to individual anatomical variability, we further computed a Population-Average Attention Map by aggregating voxel-wise intensities across all test subjects.

\section{RESULTS}
We evaluated the proposed hybrid classification framework across three clinically relevant binary classification tasks: (1) PSP vs. PD, (2) MSA vs. PD, and (3) PSP vs. MSA. For each task, we conducted an ablation study by comparing multiple input modalities (volume only, mask only, MRI only, and combinations thereof) and modeling strategies (ML only, CNN only, and CNN+ML fusion). Table~\ref{tab:merged_results} summarizes the classification performance across all three classification tasks.

\begin{table}[!htbp]
\caption{Performance comparison on held-out test data for three classification tasks using different input modalities and model types. Best values per task and metric are highlighted in bold.}
\label{tab:merged_results}
\begin{center}
\renewcommand{\arraystretch}{0.95}
\setlength{\tabcolsep}{4pt}
\resizebox{\linewidth}{!}{
\begin{tabular}{lllcccccc}
\hline
\textbf{Task} & \textbf{Model} & \textbf{Input Data} & \textbf{Sensitivity} & \textbf{Specificity} & \textbf{Youden} & \textbf{AUC} & \textbf{F1} & \textbf{Accuracy} \\
\hline
\textbf{PSP vs. PD} & ML     & Volume             & 0.82 & 0.79 & 0.62 & 0.90 & 0.84 & 0.81 \\
\cdashline{2-9}
           &     & Mask               & \textbf{0.93} & 0.84 & 0.77 & 0.94 & \textbf{0.91} & 0.89 \\
           &   CNN     & MRI                & 0.87 & 0.64 & 0.51 & 0.88 & 0.83 & 0.78 \\
           &        & MRI, Mask          & 0.91 & 0.84 & 0.75 & 0.94 & 0.90 & 0.88 \\
\cdashline{2-9}
           &        & Mask, Volume       & 0.88 & 0.82 & 0.70 & 0.95 & 0.88 & 0.85 \\
           & Hybrid & MRI, Volume        & \textbf{0.93} & 0.77 & 0.70 & 0.94 & 0.89 & 0.86 \\
           &        & MRI, Mask, Volume  & 0.87 & \textbf{0.91} & \textbf{0.78} & \textbf{0.95} & \textbf{0.91} & \textbf{0.90} \\
\hhline{=========}
\textbf{MSA vs. PD} & ML     & Volume             & 0.67 & 0.76 & 0.43 & 0.80 & 0.51 & 0.74 \\
\cdashline{2-9}
           &     & Mask               & 0.69 & 0.76 & 0.45 & 0.79 & 0.54 & 0.74 \\
           &  CNN & MRI                & 0.71 & 0.72 & 0.43 & 0.80 & 0.52 & 0.72 \\
           &        & MRI, Mask          & 0.77 & 0.75 & 0.52 & 0.83 & 0.59 & 0.76 \\
\cdashline{2-9}
           &           & Mask, Volume       & 0.67 & 0.78 & 0.44 & 0.81 & 0.53 & 0.75 \\
           &  Hybrid   & MRI, Volume        & \textbf{0.80} & 0.79 & \textbf{0.59} & \textbf{0.86} & \textbf{0.62} & \textbf{0.79} \\
           &        & MRI, Mask, Volume  & 0.67 & \textbf{0.83} & 0.49 & \textbf{0.86} & 0.57 & \textbf{0.79} \\
\hhline{=========}
\textbf{PSP vs. MSA} & ML     & Volume             & 0.67 & 0.85 & 0.51 & 0.87 & 0.65 & 0.80 \\
\cdashline{2-9}
            &     & Mask               & \textbf{0.91} & 0.67 & \textbf{0.58} & 0.82 & \textbf{0.71} & 0.73 \\
            &  CNN      & MRI                & 0.82 & 0.59 & 0.41 & 0.78 & 0.58 & 0.66 \\
            &        & MRI, Mask          & 0.85 & 0.67 & 0.52 & 0.86 & 0.65 & 0.73 \\
\cdashline{2-9}
            &  & Mask, Volume       & 0.53 & \textbf{0.95} & 0.48 & \textbf{0.92} & 0.64 & \textbf{0.83} \\
            &   Hybrid     & MRI, Volume        & 0.67 & 0.85 & 0.51 & 0.91 & 0.65 & 0.80 \\
            &        & MRI, Mask, Volume  & 0.60 & 0.92 & 0.52 & 0.88 & 0.67 & \textbf{0.83} \\
\hline
\end{tabular}}
\end{center}
\end{table}

\begin{figure}[tb]
\centering
\includegraphics[width=\textwidth]{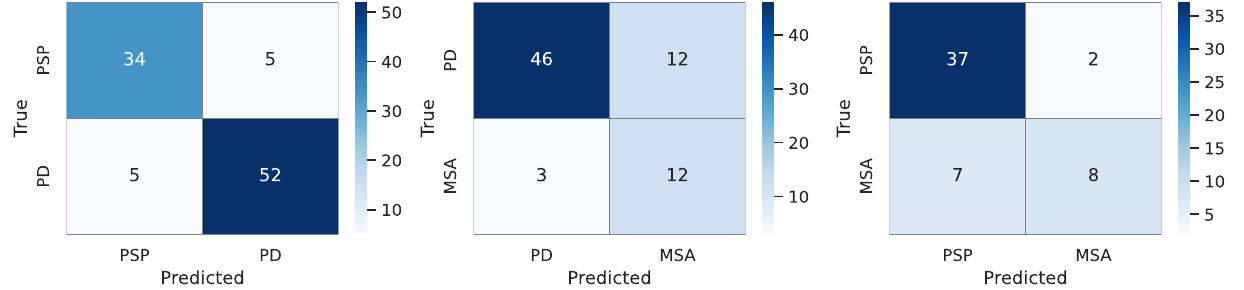}
\caption{Confusion matrices for: (left) PSP vs. PD; (middle) PD vs. MSA; (right) PSP vs. MSA.}
\label{fig:confusion}
\end{figure}

For the PSP vs. PD classification, the best overall performance was obtained by the hybrid model combining MRI, segmentation masks, and volumetric features, yielding an AUC of 0.95, F1 score of 0.91, and accuracy of 89.6\%. The hybrid model was further evaluated on the test set (38 PSP vs. 57 PD). The confusion matrix (Figure~\ref{fig:confusion}, left) indicates balanced performance across both classes, with sensitivity and specificity both above 87\%.
In the experiment of MSA vs. PD classification, we grouped the MSA-C, MSA-P, and unclassified MSA cases due to the limited number of subjects in each subgroup. Additionally, the classification was imbalanced, with 77 MSA subjects versus 286 PD subjects. Overall performance in this task was lower than PSP vs. PD. This may be due to the lower number of MSA cases used in training and the combination of MSA subtypes into one patient group. The hybrid model combining MRI and volume data reached the highest AUC of 0.86, with an F1 score of 0.62 and accuracy of 0.79. 
The final classification experiment focused on distinguishing PSP vs. MSA, which is considered the most clinically challenging task, particularly in the early stages. As in the previous experiment, we grouped all MSA cases into one patient group. We also have an imbalanced classification set, with 77 MSA subjects versus 192 PSP subjects. The best results were again achieved by the hybrid model (Mask + volume), with an AUC of 0.92 and accuracy of 0.83. 

\begin{figure}[tb]
    \centering
    \includegraphics[width=\textwidth]{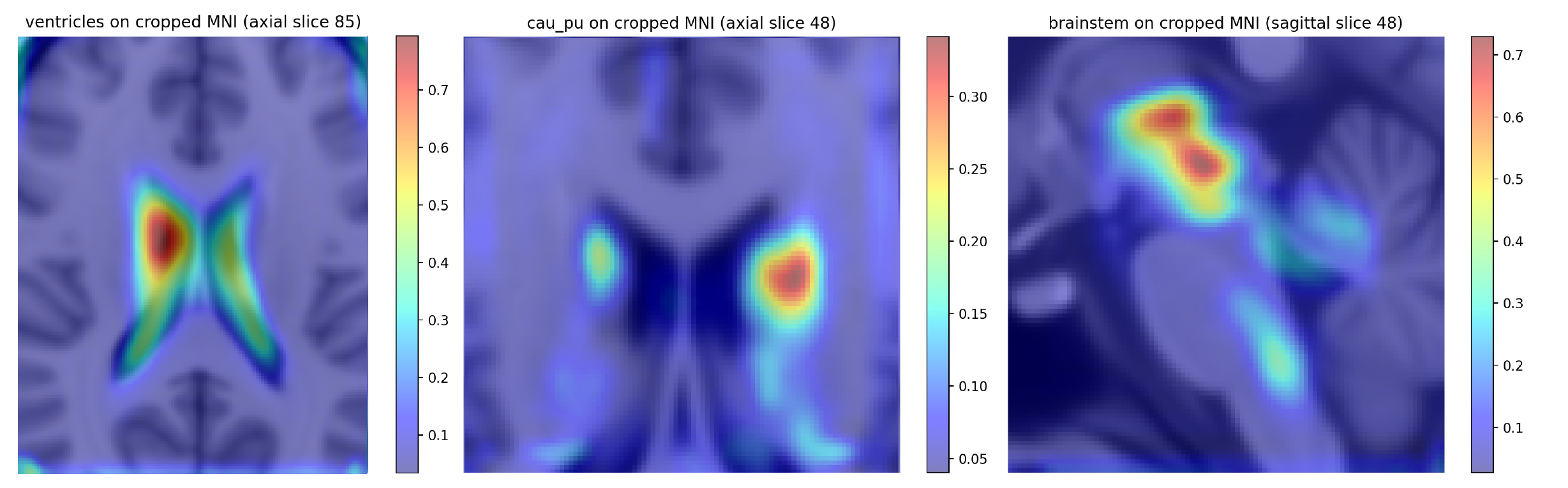}
    \caption{3D Grad-CAM attention maps for PD vs. PSP classification. The figure shows population-averaged class-discriminative attention maps computed on the held-out test set for the ventricular system (left), striatum (middle), and brainstem (right) branches. Attention maps are overlaid on the MNI (Montreal Neurological Institute) template. Warmer colors (red/yellow) indicate regions with a higher contribution to the model’s classification decision. The model consistently attends to the midbrain and ventricular regions, which are known to be associated with PSP-related neurodegeneration.}
    \label{fig:attention_map}
\end{figure}

Figure~\ref{fig:attention_map} presents the population-average attention maps computed on the held-out test set for the PD vs. PSP classification task, overlaid on a standard MNI (Montreal Neurological Institute) MRI template ~\cite{mazziotta1995probabilistic, mazziotta2001probabilistic, fonov2011unbiased}. The visualizations indicate that the model consistently focuses on anatomically and clinically meaningful regions associated with established neuropathological markers.

In the sagittal brainstem view, the network exhibits maximal activation in the dorsal midbrain tegmentum, corresponding to the anatomical substrate underlying the hummingbird sign, a hallmark of PSP-related midbrain atrophy. Additionally, the attention observed in the striatum regions, including the caudate and putamen, suggests that the model captures more subtle extranigral degeneration patterns that differentiate atypical Parkinsonism from idiopathic PD. Importantly, the activation patterns are spatially localized and confined to relevant neuroanatomical structures, indicating that the classifier relies on biologically plausible features rather than nonspecific or background signals.

\section{CONCLUSION}
This work presents a novel two-stage hybrid classification framework that integrates deep convolutional features with structure-specific volumetric data to aid in the differential diagnosis of APD using T1-weighted brain MRI. Unlike most previous approaches that rely solely on area ratios, volumetrics, or black-box CNN models, we propose a multi-branch 3D CNN architecture with spatial pyramid pooling trained on MRI and segmentation masks, and further enhancing the discriminative power by fusing learned features with ICV-corrected brain structure volumes via logistic regression. This is, to our knowledge, the first work to systematically evaluate such a multi-modal and multi-stage fusion strategy across three clinically challenging APD classification tasks: PSP vs. PD, MSA vs. PD, and PSP vs. MSA. 

 Our results show that incorporating segmentation masks and structure-specific volume data in addition to MRI images improves robustness and accuracy compared to models based on a single modality. The framework was evaluated on a large multi-site clinical dataset and tested on a held-out test set, achieving an AUC of 0.95 for PSP vs. PD classification, an AUC of 0.86 for MSA vs. PD classification, and an AUC of 0.92 for PSP vs. MSA classification, exceeding previously reported AUC values on similar classification tasks~\cite{SJOSTROM202018, Volkmann2025}. The model achieves high accuracy and AUC scores across all three classification tasks, demonstrating its potential as a reliable diagnostic aid for early APD differentiation and future clinical research. 

Beyond competitive classification performance, the proposed hybrid framework demonstrates a high degree of interpretability through 3D Grad-CAM analysis. The population-average attention maps indicate that the model consistently focuses on biologically meaningful neuroanatomical regions, including the midbrain, ventricular system, and striatum, rather than relying on diffuse or non-informative background signals. Importantly, the highlighted ventricular patterns reflect global ventricular morphology and spatial configuration instead of a single handcrafted biomarker, suggesting that the model captures complementary structural information beyond conventional indices. These findings support the potential of the proposed framework to provide clinically interpretable insights while maintaining robust discriminative capability for APD differentiation. This work provides evidence that anatomically guided, multi-modal fusion models hold strong promise for clinical decision support in early APD diagnosis. Future work will extend the framework to multi-class classification and include more extensive imaging data, including longitudinal data.


\acknowledgments
This research was supported by the Icelandic Centre for Research (RANNIS) under grant 200101-5601, the University of Iceland Doctoral Fund (Eimskip Fund), and the University of Iceland Research Fund. 

\noindent *The ASAP Neuroimaging Initiative: University of Cologne, Germany: Thilo van Eimeren, thilo.van-eimeren@uk-koeln.de, Kathrin Giehl, kathrin.giehl@uk-kolen.de, Elena Doering, elena.doering@uk-koeln.de; Turku University Hospital, Finland: Valtteri Kaasinen, valtteri.kaasinen@tyks.fi; National Research Council, Italy: Andrea Quattrone, an.quattrone@unicz.it; University of Campania, "Luigi Vanvitelli”, Italy: Alessandro Tessitore, alessandro.tessitore@unicampania.it; Universidad de Navarra, Spain: María Rodríguez Oroz, mcroroz@unav.es; University of Florida, USA: David Vaillancourt, vcourt@ufl.edu; FLENI Foundation Buenos Aires, Argentina: Julieta Arena, jarena@fleni.org.ar; University of Salerno, Italy: Marina Picillo, mpicillo@unisa.it, Paolo Barone, pbarone@unisa.it, Maria Teresa Pellecchia, m.pellecchia@unisa.it.

\bibliographystyle{spiebib}
\bibliography{report}

\end{document}